\documentclass[letterpaper]{article}
\pdfoutput=1

\usepackage{aaai}

\nocopyright

\usepackage{times}
\usepackage{helvet}
\usepackage{courier}
\usepackage{graphicx}
\usepackage{subfigure}

\usepackage{amsmath,amsthm,amsfonts}
\newcommand{\argmax}{\operatornamewithlimits{argmax}}

\usepackage[ruled, vlined]{algorithm2e}
\DontPrintSemicolon

\usepackage{multirow}

\theoremstyle{definition}

\theoremstyle{plain}
\newtheorem{lemma}{Lemma}
\newtheorem{theorem}{Theorem}

\begin{document}
%
\title{Regret-Based Multi-Agent Coordination with Uncertain Task Rewards}
\author{Feng Wu \and Nicholas R. Jennings \\
School of Electronics and Computer Science, University of
Southampton, UK \\
\texttt{\{fw6e11, nrj\}@ecs.soton.ac.uk}}

\maketitle

\begin{abstract}
\begin{quote}
Many multi-agent coordination problems can be represented as DCOPs.
Motivated by task allocation in disaster response, we extend
standard DCOP models to consider {\em uncertain} task rewards where
the outcome of completing a task depends on its current state,
which is randomly drawn from {\em unknown} distributions. The goal
of solving this problem is to find a solution for all agents that
minimizes the overall worst-case loss. This is a challenging
problem for centralized algorithms because the search space grows
exponentially with the number of agents and is nontrivial for
standard DCOP algorithms we have. To address this, we propose a
novel decentralized algorithm that incorporates Max-Sum with
iterative constraint generation to solve the problem by passing
messages among agents. By so doing, our approach scales well and
can solve instances of the task allocation problem with hundreds of
agents and tasks.
\end{quote}
\end{abstract}

\section{Introduction}
\label{sec:introduction}%

{\em Distributed constraint optimization problems} (DCOPs) are a
popular representation for many multi-agent coordination problems.
In this model, agents are represented as decision variables and the
tasks that they can be assigned are variable domains. The synergies
between agents' joint \mbox{assignment} are specified as constraint
values. Now, some tasks may require a subgroup of the team to work
together, either because a single agent has insufficient
capabilities to complete the task or the teamwork can substantially
improve performance. In either case, the constraints are the
utilities of the agents' joint behaviors. Once the DCOP model of
the problem is obtained, we can solve it efficiently using optimal
approaches such as ADOPT~\cite{ModiSTY05} and DPOP~\cite{PetcuF05}
or approximate approaches such as DSA~\cite{ZhangWXW05},
MGM~\cite{MaheswaranPT04}, and Max-Sum~\cite{FarinelliRPJ08}.

In DCOPs, the task rewards are often assumed to be completely known
to the agents. However, this can make it difficult to model
problems where the reward of completing a task depends on the task
state, which is usually unobservable and uncontrollable by the
agents. For example, in disaster response, a group of robots may be
sent out to an unknown area to search for survivors. However, the
success of the search tasks (task rewards) will depend on many
factors (task states) such as the local terrain, the weather
condition, and the degree of damage. Initially, the robots may have
very limited information about the task states, but must act
quickly because time is critical for saving lives. In such cases,
it is desirable to reason about the uncertainty of the task states
(rewards) and assign the tasks to the agents in such a way that the
worst-case loss~\footnote{Here, we focus on the worst-cast loss but
other robust optimization criterions (e.g., the maximin model)
could be applied.} (compared to the unknown optimal solution) is
minimized. The aim of this is to perform as closely as possible to
the optimal solution given the uncertain task rewards (caused by
{\em unknown} task states).

Over recent years, a significant body of research has dealt with
extending standard DCOPs to models with uncertainty. A common
method is to introduce additional random variables (uncontrollable
by the agents) to the \mbox{constraint} functions~\cite{LeauteF11}.
Another way to reason about the uncertainty is to randomly select a
constraint function from a predefined function
set~\cite{AtlasD10,StrandersFRJ11,NguyenYL12}. However, all the
aforementioned approaches require the probability distributions of
the random variables to be known~\cite{AtlasD10,LeauteF11} or the
candidate \mbox{functions} to have certain properties (e.g., be
Gaussian~\cite{StrandersFRJ11} or concave~\cite{NguyenYL12}).
Unfortunately, these assumptions are not common in our motivating
domain because the agents have no or very limited information about
the tasks as they start to respond to the crisis. Thus, the key
challenge in our domain is to find a good solution (as close to the
optimal as possible) given no or partial information about the
associated task states (linked to the rewards).

To this end, we introduce a new model for multi-agent coordination
with uncertain task rewards and propose an efficient algorithm for
computing the {\em robust} solution (minimizing the worst-case
loss) of this model. Our model, called {\em uncertain} reward DCOP
(UR-DCOP), extends the standard DCOP to include random variables
(task states), one for each constraint (task). We assume these
random variables are independent from each other (the tasks are
independent) and uncontrollable by the agents (e.g., the weather
condition). Furthermore, we assume the choice of these variables
are drawn from finite domains with unknown distributions. For each
such variable, we define a belief as a probability distribution
over its domain. Thus, minimizing the worst-case loss is equivalent
to computing the {\em minimax} regret solution in the joint belief
space. Intuitively, this process can be viewed as a game between
the agents and nature where the agents select a solution to
minimize the loss, while nature chooses a belief in the space to
maximize it.

For large UR-DCOPs, it is intractable for a centralized solver to
compute the minimax regret solution for all the agents due to the
huge joint belief and solution space. Thus, we turn to
decentralized approaches because they can exploit the interaction
structure and distribute the computation locally to each agent.
However, it is challenging to compute the minimax regret in a
decentralized manner because intuitively all the agents need to
find the worst case (a point in the belief space) before they can
minimize the loss. To address this, we borrow ideas from {\em
iterative constraint generation}, first introduced
by~\cite{Benders62} and recently adopted
by~\cite{ReganB10,ReganB11} for solving imprecise MDPs. Similar to
their approaches, we decompose the overall problem into a {\em
master} problem and a {\em subproblem} that are iteratively solved
until they converge. The main contribution of our work lies in the
development of two variations of Max-Sum~\cite{FarinelliRPJ08} to
solve the master and sub-problems by passing messages among the
agents. We adopt Max-Sum due to \mbox{its} performance and
stability on large problems (i.e., hundreds of agents). For acyclic
factor graphs, we prove our algorithm is optimal. In experiments,
we show that our method can scale up to task allocation domains
with hundreds of agents and tasks (intractable for centralized
approaches) and can outperform the state-of-the-art decentralized
approach by having much lower average regrets.

The reminder of the paper is organized as follows. First, we start
with the background and introduce our UR-DCOP model. Then, we
propose our algorithm and analyze its main properties. After that,
we present our empirical results and conclude the paper with future
work.

\section{The UR-DCOP Model}
\label{sec:model}%

Formally, a {\em distributed constraint optimization problem}
(DCOP) can be defined as a tuple $\mathcal{M} = \langle
\mathcal{I}, \mathcal{X}, \mathcal{D}, \mathcal{U} \rangle$, where:
\begin{itemize}
\item $\mathcal{I} = \{ 1, \cdots, n \}$ is a set of agents
    indexed by $1, 2, \cdots, n$;
\item $\mathcal{X} = \{ x_1, \cdots, x_n \}$ is a set of {\em
    decision variables} where $x_i$ denotes the variable
    controlled by agent $i$;
\item $\mathcal{D} = \{ D_1, \cdots, D_n \}$ is a set of finite
    domains for the decision variables where domain $D_i$ is a
    set of possible values for decision variable $x_i$;
\item $\mathcal{U} = \{ U_1, \cdots, U_m \}$ is a set of {\em
    soft} constraints where each constraint $U_j: D_{j1} \times
    \cdots \times D_{jk} \rightarrow \Re$ defines the value of
    possible assignments to subsets of decision variables where
    $U_j(x_{j1}, \cdots, x_{jk})$ is the value function for
    variables $x_{j1}, \cdots, x_{jk} \in \mathcal{X}$.
\end{itemize}
The goal of solving a DCOP is to find an assignment ${\bf x^*}$ of
values in the domains of all decision variables $x_i \in
\mathcal{X}$ that maximizes the sum of all constraints:
\begin{equation}
{\bf x^*} = \argmax_{{\bf x}} \sum_{j=1}^m U_j(x_{j1}, \cdots, x_{jk})
\end{equation}

Turning to Max-Sum, this is a decentralized message-passing
optimization approach for solving large DCOPs. To use Max-Sum, a
DCOP needs to be encoded as a special bipartite graph, called a
factor graph, where vertices represent variables $x_i$ and
functions $U_j$, and edges the dependencies between them.
Specifically, it defines two types of messages that are exchanged
between variables and functions:
\begin{itemize}
\item From variable $x_i$ to function $U_j$:
    \begin{equation}
        q_{i\rightarrow j}(x_i) = \alpha_{i\rightarrow j} +
        \sum_{k \in M(i)\backslash j} r_{k\rightarrow i}(x_i)
        \label{eq:v2f}
    \end{equation}
    where $M(i)$ denotes the set of indices of the function
    nodes connected to variable $x_i$ and $\alpha_{i\rightarrow
    j}$ is a scaler chosen such that $\sum_{x_i\in D_i}
    q_{i\rightarrow j}(x_i) = 0$.
\item From function $U_j$ to variable $x_i$:
    \begin{equation}
        r_{j\rightarrow i}(x_i) = \max_{{\bf x_j} \backslash x_i}
        \left[ U_j({\bf x_j}) + \sum_{k\in N(j)\backslash i}
        q_{k\rightarrow j}(x_k) \right]
        \label{eq:f2v}
    \end{equation}
    where $N(j)$ denotes the set of indices of the variable
    nodes connected to $U_j$ and ${\bf x_j}$ is a variable
    vector $\langle x_{j1}, \cdots, x_{jk} \rangle$.
\end{itemize}
Notice that both $q_{i\rightarrow j}(x_i)$ and $r_{j\rightarrow
i}(x_i)$ are scalar functions of variable $x_i\in D_i$.
Thus, the marginal function of each variable $x_i$ can be
calculated by:
\begin{equation}
    z_i(x_i) = \sum_{j\in M(i)} r_{j\rightarrow i}(x_i) \approx \max_{{\bf x}
    \backslash x_i} \sum_{j=1}^m U_j({\bf x_j})
\label{eq:mag}
\end{equation}
after which the assignment of $x_i$ can be selected by:
\begin{equation}
    x^*_i = \argmax_{x_i\in D_i} z_i(x_i)
\end{equation}

From this background, we now turn to the UR-DCOP model itself. In
particular, our work is mainly motivated by the task allocation
problem in disaster response scenarios~\footnote{Nevertheless, our
results are broadly applicable to other domains that have
uncertainty in task rewards.}, where a group of first responders
need to be assigned to a set of tasks in order to maximize saved
lives. This problem can be straightforwardly modeled as a DCOP
where: $\mathcal{I}$ is a set of first responders, $x_i$ is the
task assigned to responder $i$, $D_i$ is a set of tasks that can be
performed by responder $i$, and $U_j$ is the reward for the
completion of task $j$. However, in our domains, the value of $U_j$
does not only depend on the joint choice of the agents, but also on
the uncontrollable events such as fires, hurricanes, floods, or
debris flows in the disaster area. These events can be formally
abstracted as task states, which are usually unknown to the first
responders, but critical for the team performance. To model this,
we introduce UR-DCOP --- a new representation for multi-agent
coordination with uncertain task rewards.


In more detail, UR-DCOP is an extension of the original DCOP model
with two additional components:
\begin{itemize}
\item $\mathcal{E} = \{ s_1, \cdots, s_m \}$ is a set of {\em
    random variables} modeling uncontrollable stochastic
    events, e.g., fires in a building or weather in a disaster
    area, for each constraint $U_j \in \mathcal{U}$;
\item $\mathcal{S} = \{ S_1, \cdots, S_m \}$ is a set of finite
    domains, e.g., \mbox{levels} of the fire damage or
    different weather conditions, for each random variable $s_j
    \in S_j$;
\end{itemize}
The value functions are augmented to consider both decision
variables and random variables (task states), i.e., $U_j(s_j;
x_{j1}, \cdots, x_{jk})$. We assume each value function only
associates with one random variable. If multiple random variables
are associated with a value function, without loss of generality,
they can be merged into a single variable. Furthermore, we assume
the random variables are not under the control of the agents and
they are {\em independent} of the decision variables. Specifically,
their values are {\em independently} drawn from unknown probability
distributions.


Given a random variable $s_j$ in UR-DCOPs, the probability
distribution over domain $S_j$, denoted by $b_j \in \Delta(S_j)$,
is called a {\em belief} of the random variable, and ${\bf b} =
\langle b_1, \cdots, b_m \rangle$ is a {\em joint belief} of all
random variables. Similarly, a {\em \mbox{joint} assignment} of all
decision variables is denoted by ${\bf x} = \langle x_1, \cdots,
x_n \rangle$ and a {\em partial joint assignment} for the value
function $U_j$ is denoted by ${\bf x_j} = \langle x_{j1}, \cdots,
x_{jk} \rangle$. When the joint belief ${\bf b}$ is known, solving
a UR-DCOP straightforwardly involves finding an \mbox{assignment}
of all decision variables ${\bf x}$ that maximize the sum of the
{\em expected} values:
\begin{equation}
V({\bf b}, {\bf x}) = \sum_{j=1}^m \underbrace{\sum_{s_j\in S_j} b_j(s_j)
U_j(s_j, {\bf x_j})}_{U_j(b_j, {\bf x_j})}
\label{eq:exp}
\end{equation}

The key challenge in our problem is that the joint belief ${\bf b}$
is unknown. Therefore, we want to find a solution that is {\em
robust} (minimizing the worst-case loss) to the uncertainty of the
joint belief. As mentioned earlier, this objective is equivalent to
the {\em minimax regret} given the belief space $\mathcal{B}$:
\begin{equation}
V_{\mbox{\it regret}}({\bf x}) = \min_{{\bf x'}}\underbrace{
\max_{{\bf b}\in\mathcal{B}} \underbrace{\max_{{\bf x^*}}
[ V({\bf b}, {\bf x^*}) - V({\bf b}, {\bf x'}) ]}_{R_1({\bf x'}, {\bf b})}
}_{R_2({\bf x'})}
\label{eq:regret}
\end{equation}
where ${\bf x^*}$ is the optimal solution given belief ${\bf b}$.
$R_1({\bf x'}, {\bf b})$ is the regret or loss of solution ${\bf
x}$ relative to ${\bf b}$, i.e., the difference in expected value
between ${\bf x}$ and the optimal solution ${\bf x^*}$ under belief
${\bf b}$. $R_2({\bf x'})$ is the maximum regret of ${\bf x}$ with
respect to the feasible belief space. Thus, the value of minimax
regret, $V_{\mbox{\it regret}}({\bf x})$, minimizes the worst-case
loss over all possible belief points.

As mentioned, first responders usually have very limited
information about the response tasks when the disaster happens and
there is significant uncertainty in the \mbox{environment}. In such
cases, {\em minimax regret} minimizes the difference between the
optimal value $V({\bf b}, {\bf x^*})$ and the actual value $V({\bf
b}, {\bf x})$ achieved by the current solution ${\bf x}$ in all
possible beliefs ${\bf b}\in \mathcal{B}$. Thus, it is a good
solution for the first responders given the limited information.

\section{Solving UR-DCOPs}
\label{sec:algorithm}%

Generally, to compute the minimax regret in
Equation~\ref{eq:regret}, we first need to compute the optimal
solution ${\bf x^*}$ given a belief point ${\bf b}$ and the current
solution of agents ${\bf x'}$. Then, the whole belief space
$\mathcal{B}$ is searched to find the worst-case belief ${\bf b}$.
After that, we need to find the assignment ${\bf x}$ that minimizes
the regret. On the one hand, it cannot be solved by standard DCOP
algorithms. On the other hand, given a number of agents, it is very
challenging for centralized algorithms to compute the minimax
regret because the search space blows up exponentially with the
number of agents.

Following the ideas of {\em Iterative Constraint Generation} (ICG),
two optimizations are alternatively solved at each iteration: the
{\em master problem} and the {\em subproblem}. In more detail, the
{\em master} problem solves a relaxation of
Equation~\ref{eq:regret} by considering only a subset of all
possible $\langle {\bf b}, {\bf x^*} \rangle$ pairs $\mathcal{G}$:
\begin{equation}
  \begin{array}{ll}
    \min\nolimits_{{\bf x}, \delta} & \delta \\
	\mbox{s. t.} & \forall \langle {\bf b}, {\bf x^*} \rangle
    \in \mathcal{G}, V({\bf b}, {\bf x^*}) - V({\bf b}, {\bf x}) \leq \delta
  \end{array}
  \label{eq:master}
\end{equation}
Initially, this set can be arbitrary (e.g., empty or randomly
generated). By giving $\mathcal{G}$, the master problem tries to
minimize the loss for the worst case derived from $\mathcal{G}$.

The {\em subproblem} generates the maximally violated constraint
relative to ${\bf x}$, the solution of the current master problem.
More precisely, a new $\langle {\bf b}, {\bf x^*} \rangle$ pair is
found by the subproblem. This pair is called a {\em witness} point
because it indicates that the current ${\bf x}$ is not the best
solution in terms of the minimax regret. In more detail, a program
is solved to determine the witness $\langle {\bf b}, {\bf x^*}
\rangle$ for the current solution ${\bf x}$:
\begin{equation}
  \begin{array}{ll}
    \max_{{\bf b}, {\bf x^*}, \delta'} & \delta' \\
    \mbox{s. t.} & V({\bf b}, {\bf x^*}) - V({\bf b}, {\bf x}) \geq \delta'
  \end{array}
  \label{eq:sub}
\end{equation}
If $\delta' = \delta$ then the constraint for $\langle {\bf b},
{\bf x^*} \rangle$ in Equation~\ref{eq:sub} is satisfied at the
current solution ${\bf x}$, and indeed all unexpressed constraints
must be satisfied as well. Otherwise, $\delta' > \delta$, implying
that the constraint for $\langle {\bf b}, {\bf x^*} \rangle$ is
violated in the current relaxation. Thus, it is added to
$\mathcal{G}$ and the master problem is solved again to compute a
new ${\bf x}$. This process repeats until no new witness point can
be found by the subproblem and the master problem terminates with
the best solution ${\bf x}$.

Based on the ideas of ICG, we propose {\em Iterative Constraint
Generation Max-Sum} (ICG-Max-Sum) to solve UR-DCOPs. Similar to
standard Max-Sum, our algorithm starts with encoding UR-DCOPs into
a factor graph. Then, two Max-Sum algorithms are iteratively
executed to solve the master and sub-problems. In the master
problem, we run a Max-Sum to compute the current minimax solution
${\bf x}$ and minimax regret $\delta$ given the witness set
$\mathcal{G}$. In the subproblem, we run another Max-Sum to
generate a new witness point $\langle {\bf b}, {\bf x^*} \rangle$
and the corresponding minimax regret $\delta'$ given the current
solution ${\bf x}$. Then, $\delta$ and $\delta'$ are compared by
each node in the factor graph: If $\delta>\delta'$, the newly
generated witness point $\langle {\bf b}, {\bf x^*} \rangle$ is
added to $\mathcal{G}$; otherwise it terminates and returns the
current minimax solution ${\bf x}$. These processes repeat until
all nodes in the factor graph are terminated. Notice that in our
algorithm the solutions $x_i \in {\bf x}$ and $x^*_i \in {\bf x^*}$
are computed and stored locally by variable $i$ and belief $b_j \in
{\bf b}$ is computed and stored locally by function $j$. The main
procedures are shown in Algorithm~\ref{alg:icg-maxsum}.

\begin{algorithm}[!t]
  \caption{Iterative Constraint Generation Max-Sum}
  \KwIn{$\mathcal{M}$: The UR-DCOP Model}
  Create a factor graph based on $\mathcal{M}$ \;
  Initialize the witness set $\mathcal{G} \gets \emptyset$ \;
  \Repeat{all nodes in the graph are terminated.}{
    \tcp{The Master Problem}
    Run Max-Sum on the factor graph with $\mathcal{G}$ \;
    Compute the current minimax solution ${\bf x}$ \;
    Save each $x_i \in {\bf x}$ in variable node $i$ \;
    Compute the minimax regret $\delta$ \;
    \tcp{The Subproblem}
    Run Max-Sum on the factor graph with ${\bf x}$ \;
    Compute the witness point $\langle {\bf b}, {\bf x^*} \rangle$ \;
    Compute the minimax regret $\delta'$ \;
    \tcp{$\mathcal{G} \gets \mathcal{G} \cup
    \{ \langle {\bf b}, {\bf x^*} \rangle \}$ if $\delta' > \delta$}
    \ForEach{variable node $i$}{
        \If{$\delta' > \delta$}{
            Save $x^*_i \in {\bf x^*}$ in variable node $i$ \;
        }
        {\bf else} Terminate variable node $i$ \;
    }
    \ForEach{function node $j$}{
        \If{$\delta' > \delta$}{
            Save $b_j \in {\bf b}$ in function node $j$ \;
        }
        {\bf else} Terminate function node $j$ \;
    }
  }
  \Return the current minimax solution ${\bf x}$ \;
  \label{alg:icg-maxsum}
\end{algorithm}

\subsection{The Master Problem}

The {\em master} problem of Equation~\ref{eq:master}, given the
witness set $\mathcal{G}$, can be equivalently written as:
\begin{equation}
{\bf x} = \arg\underbrace{\min_{{\bf x'}} \max_{\langle {\bf
b}, {\bf x^*}\rangle \in \mathcal{G}} \sum_{j=1}^m
[ U_j(b_j, {\bf x^*_j}) - U_j(b_j, {\bf x'_j}) ]}_{\delta}
\end{equation}
Note that the witness set $\mathcal{G}$ is known and the choice of
$\langle b_j, {\bf x^*_j} \rangle$ can be computed locally by
function node $j$ in Max-Sum because $b_j$ is independent from
other belief points and ${\bf x^*_j}$ is only related to the
variable nodes it connects. To do this, we consider the problem of
minimizing a vector of regret functions for each witness point in
$\mathcal{G}$:
\begin{equation}
\mathbb{V}({\bf x}) = \left[ \tilde{V}({\bf x}, \langle {\bf b}, {\bf x^*} \rangle_1),
\cdots, \tilde{V}({\bf x}, \langle {\bf b}, {\bf x^*} \rangle_{|\mathcal{G}|}) \right]
\end{equation}
where $\tilde{V}({\bf x}, \langle {\bf b}, {\bf x^*} \rangle_g) =
V({\bf b}, {\bf x^*}) - V({\bf b}, {\bf x})$ and $\langle {\bf b},
{\bf x^*} \rangle_g$ is the $g$th element in $\mathcal{G}$.
Accordingly, instead of $q_{j\rightarrow i}(x_i)$ and
$r_{i\rightarrow j}(x_i)$ being scalar functions of $x_i$, these
messages now map the domain of $x_i$ to a set of regret vectors:
$\forall x_i\in D_i, q_{j\rightarrow i}(x_i) = [ q_1, \cdots,
q_{|\mathcal{G}|} ], r_{i\rightarrow j}(x_i) = [ r_1, \cdots,
r_{|\mathcal{G}|} ]$.

To compute these messages, the two key operators required by
Max-Sum (Equations~\ref{eq:v2f} and~\ref{eq:f2v}) need to be
redefined.
In more detail, adding two messages is defined to add each
corresponding element in the two vectors: $q^1_{j\rightarrow
i}(x_i) + q^2_{j\rightarrow i}(x_i) = [ q^1_1 + q^2_1, \cdots,
q^1_{|\mathcal{G}|} + q^2_{|\mathcal{G}|} ]$ and $r^1_{i\rightarrow
j}(x_i) + r^2_{i\rightarrow j}(x_i) = [ r^1_1 + r^2_1, \cdots,
r^1_{|\mathcal{G}|} + r^2_{|\mathcal{G}|}]$. For
Equation~\ref{eq:f2v}, we need to minimize the regret of function
node $j$ with respect to its neighboring variables ${\bf x_j}$ as:
\begin{equation}
r_{j\rightarrow i}(x_i) = \mathbb{U}_j({\bf \tilde{x}_j}) +
\sum_{k\in N(j)\backslash i} q_{k\rightarrow j}(\tilde{x}_k)
\label{eq:msg}
\end{equation}
where $\mathbb{U}_j({\bf \tilde{x}_j}) = [\tilde{U}_j({\bf
\tilde{x}_j}, \langle b_j, {\bf x^*_j} \rangle_1), \cdots,
\tilde{U}_j({\bf \tilde{x}_j}, \langle b_j, {\bf x^*_j}
\rangle_{|\mathcal{G}|}]$, $\tilde{U}_j({\bf \tilde{x}_j}, \langle
b_j, {\bf x^*_j} \rangle_g) = U_j(b_j, {\bf x^*_j}) - U_j(b_j, {\bf
x'_j})$, and
\begin{equation}
\begin{split}
{\bf \tilde{x}_j} = \arg\min_{{\bf x_j}\backslash x_i}
\max_{\langle b_j, {\bf x^*_j} \rangle_g \in \mathcal{G}}
[ \tilde{U}_j({\bf x_j}, \langle b_j, {\bf x^*_j} \rangle_g) + \\
\sum_{k\in N(j)\backslash i} q_{k\rightarrow j}(x_k, g) ]
\end{split}
\label{eq:msgarg}
\end{equation}

At the end of the message-passing phase, each variable $x_i$
computes its marginal function $z_i(x_i)$ according to
Equation~\ref{eq:mag}. Obviously, the value of the marginal
function is also a vector: $z_i(x_i) = [ z_1, \cdots,
z_{|\mathcal{G}|} ]$. The best assignment of the variable $x_i$ can
be computed by:
\begin{equation}
x_i = \arg\min_{x'_i\in D_i} \max_{g} z_i(x'_i, g)
\end{equation}
where $g$ is an index for the vector. After that, the minimax
regret $\delta$ can be computed by propagating values in a (any)
pre-defined tree structure of the factor graph: (1) Each variable
node send its assignment to its neighboring nodes; (2) On received
all the assignments from its neighboring nodes, each function node
computes the regret value and sent the message to its neighboring
nodes; (3) Each node propagates the regret values until all the
regret values are computed and received by all the nodes. Then,
$\delta$ can be computed by adding all the $m$ messages in each
node.

\begin{figure}[t]
  \centering\small
  \subfigure[b][$\tilde{V}({\bf x}, \langle {\bf b}, {\bf x^*} \rangle)$]{
    \begin{tabular}{c|cc}
         &  G1   &  G2    \\ \hline
      AC & -57   &  64    \\
      AD & -96   & -162   \\
      BC &  54   &  72    \\
      BD & -4    &  55    \\
    \end{tabular}}
  \subfigure[b][{\scriptsize Factor Graph}]{
      \raisebox{-23pt}[0pt][25pt]{
        \includegraphics[width=0.2\linewidth]{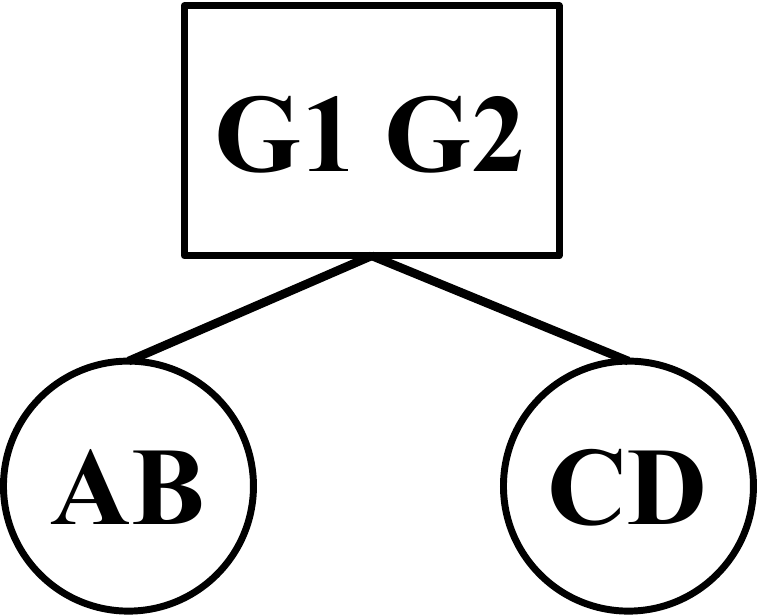}}}
  \subfigure[b][$z_i(x_i, \langle {\bf b}, {\bf x^*} \rangle)$]{
    \begin{tabular}{c|cc}
        &  G1   &  G2    \\ \hline
      A & -96   & -162   \\
      B & -4    &  55    \\ \hline
      C & -57   &  64    \\
      D & -96   & -162   \\
    \end{tabular}} \vspace{-5pt}
  \caption{Example of the Master Problem.}
  \label{fig:example}
\end{figure}

An example of the master problem with randomly generated
$\mathbb{V}$ is shown in Figure~\ref{fig:example}. In this example,
there are two variables with the domain $\{ A, B\}$ and $\{ C, D
\}$ respectively and the witness set $\mathcal{G}$ is $\{ G1, G2
\}$. Clearly, the minimax solution is $AD$ and the minimax regret
is $-96$ \mbox{since} we have $\min\{ \max\{-57, 64\}$, $\max\{-96,
-162\}$, $\max\{54, 72\}$, $ \max\{-4, 55\}\}$ = $-96$. For our
Max-Sum, according to Equation~\ref{eq:msg}, the message $r_1(A)$ =
$\mathbb{V}(AD)$ since $AD$ = $\arg\min_{AD, AC}\{ \max\{ -57, 64
\}, \max\{ -96, -162 \} \}$. Similarly, we have the messages:
$r_1(B)$ = $\mathbb{V}(BD)$, $r_1(C)$ = $\mathbb{V}(AC)$, and
$r_1(D)$ = $\mathbb{V}(AD)$. After the message-passing phase, the
marginal functions $z_1(A)$ = $\mathbb{V}(AD)$, $z_1(B)$ =
$\mathbb{V}(BD)$, $z_2(C)$ = $\mathbb{V}(AC)$, $z_2(D)$ =
$\mathbb{V}(AD)$. Therefore, the best assignments of each variable
are $x_1$ = $\arg\min_{A, B}\{ \max\{ -96, -162, \}, \max\{ -4,
55\} \}$ = $A$ and $x_2$ = $\arg\max_{C, D}\{ \max\{ -57, 64 \},
\max\{ -96, -162 \} \}$ = $D$. The joint solution is $AD$ and the
minimax regret is $-96$, which are equal to the minimax solution
and regret that we computed earlier according to the definition.

\subsection{The Subproblem}

The {\em subproblem} in Equation~\ref{eq:sub} given the current
solution ${\bf x}$ can be written as:
\[
\langle {\bf b}, {\bf x^*}\rangle = \arg\underbrace{
\max_{{\bf b}\in\mathcal{B}} \max_{{\bf x'^*}} \sum_{j=1}^m
[ U_j(b_j, {\bf x'^*_j}) - U_j(b_j, {\bf x_j}) ]}_{\delta'}.
\]

Since each belief $b_j$ is independent from each other and from the
decision variables, the calculation of each belief can be moved
inside the utility function, shown as:
\[
\langle {\bf b}, {\bf x^*}\rangle =
\arg\underbrace{\max_{{\bf x'^*}} \sum_{j=1}^m \left\{\max_{b_j} [
U_j(b_j, {\bf x'^*_j}) - U_j(b_j, {\bf x_j}) ]\right\}}_{\delta'}
\]
Thus, we can define a new utility function as:
\begin{equation}
U_j({\bf x'^*_j}) = \max_{b_j} [ U_j(b_j, {\bf x'^*_j})
- U_j(b_j, {\bf x_j}) ]
\label{eq:subutil}
\end{equation}
and rely on a linear program to compute the utility:
\begin{equation}
\begin{array}{ll}
\max_{b_j} & U_j(b_j, {\bf x'^*_j})
- U_j(b_j, {\bf x_j}) \\
\mbox{s. t.} &
\forall s_j\in S_j, b_j(s_j) \geq 0 \\
& \sum_{s_j\in S_j} b_j(s_j) = 1
\end{array}
\label{eq:lp}
\end{equation}
This can be done locally and thereby the subproblem can be solved
by standard DCOP algorithms. For Max-Sum, we need to implement a
linear program (Equation~\ref{eq:lp}) in each function node to
compute the belief $b_j$ when $U_j({\bf x_j})$ is called in
Equation~\ref{eq:f2v}. Once the optimal solution ${\bf x^*}$ is
found, we can propagate ${\bf x}$ and ${\bf x^*}$ to the function
nodes and apply Equation~\ref{eq:lp} for each function node $j$ to
compute the belief ${\bf b}$. Similar to the master problem, the
minimax regret $\delta'$ can be computed by value propagation in
the factor graph.

\subsection{Analysis and Discussion}

Inherited from Max-Sum, the optimality of our algorithm depends on
the structure of the factor graph. Specifically, for an acyclic
factor graph, it is known that Max-Sum can converge to the optimal
solution of the DCOPs in finite rounds of message
passing~\cite{FarinelliRPJ08}.

\begin{lemma}
The master problems in ICG-Max-Sum will converge to the optimal
solution for acyclic factor graphs. \label{lem:master}
\end{lemma}\vspace{-5pt}
\begin{proof}[Proof (Sketch)]
The messages (vectors) in the master problems represent the regret
values of all the witness points in $|\mathcal{G}|$. The {\em sum}
operator adds up all the regret components for each witness point,
$U_j(b_j, {\bf x^*_j}) - U_j(b_j, {\bf x_j})$, sent from its
neighboring nodes. The {\em max} operator selects the
\mbox{current} minimax solution ${\bf \tilde{x}_j}$ and sends out
the corresponding regret values. Specifically, this operator is
over matrices $[m_{ij}]$ with the row $i$ indexed by witness points
and column $j$ by assignments. It compares two matrices and outputs
the one with smaller $\min_j\max_i[m_{ij}]$ value. This operator is
associative and commutative with an identity element (matrix)
$[\infty]$ (i.e., the algebra is a commutative semi-ring). Thus,
since Max-Sum is a GDL algorithm~\cite{AjiM00}, the results hold
for acyclic factor graphs.
\end{proof}

\begin{lemma}
The subproblems in ICG-Max-Sum will converge to the optimal
solution for acyclic factor graphs. \label{lem:sub}
\end{lemma}\vspace{-5pt}
\begin{proof}[Proof (Sketch)]
The subproblems are standard DCOPs given the utility function
(Equation~\ref{eq:subutil}) that can be computed locally by each
function node. Thus, Max-Sum will converge to the optimal solution
for acyclic factor graphs.
\end{proof}

\begin{theorem}
ICG-Max-Sum will converge to the optimal minimax solution for
acyclic factor graphs. \label{thm:converge}
\end{theorem}\vspace{-5pt}
\begin{proof}[Proof (Sketch)]
According to Lemmas~\ref{lem:master} and~\ref{lem:sub}, the master
problems and subproblems are optimal for acyclic factor graphs.
Thus, this theorem can be proved by showing that the subproblem
will enumerate all $\langle {\bf b}, {\bf x^*} \rangle$ pairs if
${\bf x}$ is not the minimax optimal solution. This is equivalent
to proving that in the subproblem, $\delta' > \delta$ is always
true and the new witness $\langle {\bf b}, {\bf x^*} \rangle
\not\in \mathcal{G}$ if ${\bf x} \neq \bar{{\bf x}}$ where
$\bar{\bf x}$ is the minimax optimal solution. Suppose $\delta' =
\delta$ and ${\bf x} \neq \bar{{\bf x}}$, then we have
\[
\begin{array}{lll}
\delta' & = & \max_{\langle {\bf b}, {\bf x^*} \rangle}
[V({\bf b}, {\bf x^*}) - V({\bf b}, {\bf x})] \\
& > & \min_{{\bf x'}} \max_{\langle {\bf b}, {\bf x^*} \rangle}
[V({\bf b}, {\bf x^*}) - V({\bf b}, {\bf x'})] \\
& = & \max_{\langle {\bf b}, {\bf x^*} \rangle}
[V({\bf b}, {\bf x^*}) - V({\bf b}, \bar{{\bf x}})] \Longrightarrow \\
\delta & = & \max_{\langle {\bf b}, {\bf x^*} \rangle \in \mathcal{G}}
[V({\bf b}, {\bf x^*}) - V({\bf b}, {\bf x})] = \delta' \\
& > & \max_{\langle {\bf b}, {\bf x^*} \rangle}
[V({\bf b}, {\bf x^*}) - V({\bf b}, \bar{{\bf x}})].
\end{array}
\]
Because $\mathcal{G}$ is only a subset of the whole space, we have
\[
\max_{\langle {\bf b}, {\bf x^*} \rangle \in \mathcal{G}}
[V({\bf b}, {\bf x^*}) - V({\bf b}, {\bf x})] >
\max_{\langle {\bf b}, {\bf x^*} \rangle \in \mathcal{G}}
[V({\bf b}, {\bf x^*}) - V({\bf b}, \bar{{\bf x}})].
\]
Then, the current solution ${\bf x}$ computed by the master problem
is ${\bf x} = \arg\min_{{\bf x'}} [ \max_{\langle {\bf b}, {\bf
x^*} \rangle \in \mathcal{G}} [V({\bf b}, {\bf x^*}) - V({\bf b},
{\bf x'})] ] = \bar{{\bf x}}$. This is contradictory to the
assumption ${\bf x} \neq \bar{{\bf x}}$. Furthermore, in the
subproblem, the newly generated witness point must not be in
$\mathcal{G}$, otherwise $\delta' = \delta$ due to the same ${\bf
x}$ and $\langle {\bf b}, {\bf x^*} \rangle$ being in both problems
because we have $\delta' = \max_{\langle {\bf b}, {\bf x^*}
\rangle} [V({\bf b}, {\bf x^*}) - V({\bf b}, {\bf x})] =
\max_{\langle {\bf b}, {\bf x^*} \rangle \in \mathcal{G}} [V({\bf
b}, {\bf x^*}) - V({\bf b}, {\bf x})] = \delta$.

The algorithm will converge to the minimax optimal solution
$\bar{{\bf x}}$ once all witness points $\langle {\bf b}, {\bf x^*}
\rangle$ are enumerated and added to $\mathcal{G}$ by the
subproblems. Thus, the results hold.
\end{proof}

When the factor graph is cyclic, the straightforward application of
Max-Sum is not guaranteed to converge optimally. However, in
practice, Max-Sum can produce high quality solutions even on cyclic
graphs~\cite{FarinelliRPJ08}. Moreover, it is straightforward for
our algorithm to incorporate other approximate techniques to
generate acyclic factor graphs by pruning edges~\cite{RogersFSJ11}
or adding directions~\cite{ZivanP12} in the cyclic factor graphs.
The discussion of them is beyond the scope of this paper.

For the computation and communication complexity, the subproblem
uses the standard Max-Sum except that a linear program is solved
each time when the utility function is called in
Equation~\ref{eq:f2v}. For the master problem, according to
Equation~\ref{eq:msgarg}, the computation is exponential only in
the number of variables in the scope of $U_j$ (similar to standard
Max-Sum) but {\em linear} in the number of witness points in
$\mathcal{G}$. The messages in the master problem are vectors with
the length of $|\mathcal{G}|$ while the messages in the
sub-problems are normal Max-Sum messages. In experiments, we
observed $\mathcal{G}$ is usually very small ($<$10) for the tested
problems.

\section{Empirical Evaluation}
\label{sec:experiments}%

We tested the performance of our algorithm on randomly generated
instances of the task allocation problem that we used to motivate
our work. We first generate a problem with a set of tasks $T$ and
agents $A$. Each task has a set of states $S$, from which we
randomly select one as its task state. We then create a random
graph with links among agents and tasks. Each link represents the
fact that the agent can perform the task. The utility function (a
Gaussian function whose mean and variance are randomly generated
between the ranges of 80 to 100 and 0 to 80 respectively) of each
task depends on all connected agents and its current state. In the
experiments, we set $|A|$$=$$|T|/2$ so that not all tasks can be
performed at the same time because a task requires at least one
agent. Thus, the agents must make a good choice to maximize the
team performance. We run the algorithms to solve the problem and
output its solution. Since the real states of tasks are hidden, we
want the solution to be as close to the optimal solution as
possible. To empirically evaluate this, we randomize the task
states for 100 runs and compute the average regret value (the
difference between the optimal value and the solution value given
the states) for each algorithm. For acyclic graphs, the optimal
value is computed by Max-Sum given the underlying task states.

To date, none of the existing algorithms in the literature can
solve our model so a directed comparison is not possible.
Therefore, to test the scalability and solution quality of our
algorithm, we compared it with two baseline approaches: a
centralized method based on ICG (Equations~\ref{eq:master}
and~\ref{eq:sub}) and a decentralized method based on
DSA~\cite{ZhangWXW05}. Specifically, the two operators $\max_{\bf
x^*}$ and $\min_{\bf x'}$ are alternatively solved by DSA and a
linear program is used to solve the operator $\max_{\bf
b\in\mathcal{B}}$ in Equation~\ref{eq:regret}. We ran our
experiments on a machine with a 2.66GHZ Intel Core 2 Duo and 4GB
memory. All the algorithms including Max-Sum were implemented in
Java 1.6, and the linear programs are solved by CPLEX 12.4. For
each instance, we used the same random seeds so that all the
randomized problems and task states are identical for all
algorithms.

\begin{table}[!t]
  \centering\small
  \caption{Runtime Results of ICG vs. ICG-Max-Sum}
  \vspace{4pt}
  \begin{tabular}{|c|c|c|c|c|}
    \hline
    ${\bf |A|}$ & ${\bf |T|}$ & ${\bf |S|}$ &
    {\bf ICG} & {\bf ICG-Max-Sum}  \\ \hline
    \hline
    7  & 14 & 25 & 31.41s & 5.75s \\ \hline
    7  & 14 & 50 & 53.22s & 9.07s \\ \hline
    10 & 20 & 25 & $>$ 2h & 3.53s \\ \hline
    10 & 20 & 50 & $>$ 2h & 4.38s \\ \hline
    100 & 200 & 25 & $>$ 12h & 117.04s \\ \hline
    100 & 200 & 50 & $>$ 12h & 191.93s \\ \hline
  \end{tabular}
  \label{tab:runtime}
\end{table}

\begin{table}[!t]
  \centering\small
  \caption{Regret Results of DSA vs. ICG-Max-Sum}
  \vspace{4pt}
  \begin{tabular}{|c|c|c||c|c|}
    \hline
    ${\bf |A|}$ & ${\bf |T|}$ & ${\bf |S|}$ &
    {\bf DSA} & {\bf ICG-Max-Sum} \\ \hline
    \hline
    \multicolumn{5}{|c|}{Acyclic Graphs} \\ \hline
    20 & 40 & 25 & 5973.65  & 355.49   \\ \hline
    20 & 40 & 50 & 2567.29  & 446.51   \\ \hline
    100 & 200 & 25 & 23339.31 & 475.67 \\ \hline
    100 & 200 & 50 & 17528.18 & 965.22 \\ \hline
    \hline
    \multicolumn{5}{|c|}{Cyclic Graphs} \\ \hline
    5 & 10 & 25 & 797.83 & 48.91    \\ \hline
    5 & 10 & 50 & 848.94 & 926.02   \\ \hline
    7 & 14 & 25 & 2093.86 & 1247.07 \\ \hline
    7 & 14 & 50 & 1646.29 & 1651.56 \\ \hline
  \end{tabular}
  \label{tab:regret}
\end{table}

In more details, Table~\ref{tab:runtime} shows the runtime of
(centralized) ICG and ICG-Max-Sum. We can see from the table that
the runtime of ICG increased dramatically with the problem size and
ran out of time ($>$2h) for problems with $|A|$=10, while
ICG-Max-Sum took less than 5 seconds to solve the same problems.
For ICG-Max-Sum, the problems $|A|$=10 needed less time than
problems $|A|$=7 because they took fewer iterations to converge.
Clearly from the table, large UR-DCOPs are intractable for
centralized ICG. The reason for the stability of ICG-Max-Sum is
that it can exploit the interaction structures of the task
allocation problems (tasks usually require few agents to
coordinate).

Table~\ref{tab:regret} shows the average regrets (the mean)
achieved by DSA and ICG-Max-Sum. Note that we feeded both methods
with exactly the same problems and executed them until they
converged. From acyclic graphs, we can see that ICG-Max-Sum
produced much lower regrets than DSA in the tested instances. For
large problems, DSA hardly converged to any meaningful results
(behaving like random solutions) because the errors in the
$\max_{\bf x^*}$ and $\min_{\bf x'}$ steps could accumulate over
time. Moreover, we observed that DSA took more time (one order of
magnitude) than ICG-Max-Sum for large domains because it needs to
solve $\max_{\bf x^*}$ for each $\min_{\bf x'}$ step. This confirms
the advantage of ICG-Max-Sum for solving large UR-DCOPs with lower
average regret and faster runtime. For cyclic graphs, we only
tested on small instances because it is hard to obtain optimal
solutions (the ground truth) for large problems (Max-Sum is not
optimal on cyclic graphs). In these tests, we can see from the
table that ICG-Max-Sum performed similar to \mbox{DSA} (or a little
worse especially for problems with more task states). Thus, the
aforementioned approximation techniques is useful to run Max-Sum on
cyclic graphs and bound the errors.

\section{Conclusions}
\label{sec:conclusion}%

We have presented the ICG-Max-Sum algorithm to find robust
solutions for UR-DCOPs. Specifically, we assume the distributions
of the task states are unknown and we use minimax regret to
evaluate the worst-case loss. Building on the ideas of iterative
constraint generation, we proposed a decentralized algorithm that
can compute the minimax regret and solution using Max-Sum. Similar
to Max-Sum, it can exploit the interaction structures among agents
and scale up to problems with large number of agents. We
empirically evaluated the performance of our algorithms on our
motivating task allocation domains. The experimental results show
that our algorithm has better scalability (e.g., 100 agents and 200
tasks) than the centralized method (ICG) and outperforms the
state-of-the-art decentralized method (DSA) with much lower average
regrets.

There are several future research directions that follow on from
this work. First, it would be useful to bound the message size in
the master problem when communication is very costly. Second, it
would be interesting to extend our work to domains where the tasks
are not completely independent.


\clearpage
\bibliographystyle{aaai}
\bibliography{aaai2013}

\end{document}